\documentclass[a4paper, 10pt, conference]{ieeeconf}      

\IEEEoverridecommandlockouts                              

\usepackage{graphicx}
\usepackage{subcaption}
\usepackage{graphics} 
\usepackage{svg} 
\usepackage{hyperref} 
\usepackage{url}
\usepackage{float}
\usepackage{listings}
\usepackage{xcolor}
\usepackage{tikz}
\usepackage{textcomp}
\lstdefinelanguage{YAML}{
  keywords={true,false,null,y,n},
  keywordstyle=\color{blue}\bfseries,
  basicstyle=\ttfamily\small,
  comment=[l]{\#},
  commentstyle=\color{gray}\ttfamily,
  stringstyle=\color{red}\ttfamily,
  moredelim=[l][\color{orange}]{:},
  morestring=[b]',
  morestring=[b]"
}

\title{\LARGE \bf
Fast and Realistic Automated Scenario Simulations and Reporting for an Autonomous Racing Stack
}

\author{Giovanni Lambertini$^{1}$, Matteo Pini$^{2}$, Eugenio Mascaro$^{1}$, Francesco Moretti$^{3}$, Ayoub Raji$^{1}$, Marko Bertogna$^{1}$
\thanks{$^{1}$G. Lambertini, E. Mascaro, A. Raji and M. Bertogna with the Department of Physics, Informatics and Mathematics, University of Modena and Reggio Emilia, 41121 Modena, Italy. {\tt\small firstname.lastname@unimore.it}}%
\thanks{$^{2}$M. Pini with the Department of Physics, Informatics and Mathematics, University of Modena and Reggio Emilia, 41121 Modena, Italy. {\tt\small 333250@studenti.unimore.it}}%
\thanks{$^{3}$F. Moretti with the Department of Engineering "Enzo Ferrari", University of Modena and Reggio Emilia, 41121 Modena, Italy. {\tt\small firstname.lastname@unimore.it}}%
}

\newcommand\copyrighttext{%
\footnotesize \copyright 2025 IEEE. Personal use of this material is permitted. Permission from IEEE must be obtained for all other uses, in any current or future media, including reprinting/republishing this material for advertising or promotional purposes, creating new collective works, for resale or
redistribution to servers or lists, or reuse of any copyrighted component of this work in other works.}

\newcommand\copyrightnotice{%
\begin{tikzpicture}[remember picture,overlay]
\node[anchor=north,yshift=-1cm] at (current page.north) {\parbox{\dimexpr\textwidth-\fboxsep-\fboxrule\relax}{\centering \copyrighttext}};
\end{tikzpicture}%
}

\begin{document}

\maketitle
\copyrightnotice
\thispagestyle{empty}
\pagestyle{empty}

\begin{abstract}

In this paper, we describe the automated simulation and reporting pipeline implemented for our autonomous racing stack, ur.autopilot. The backbone of the simulation is based on a high-fidelity model of the vehicle interfaced as a Functional Mockup Unit (FMU). The pipeline can execute the software stack and the simulation up to three times faster than real-time, locally or on GitHub for Continuous Integration/Continuous Delivery (CI/CD). As the most important input of the pipeline, there is a set of running scenarios. Each scenario allows the initialization of the ego vehicle in different initial conditions (position and speed), as well as the initialization of any other configuration of the stack. This functionality is essential to validate efficiently critical modules, like the one responsible for high‑speed overtaking maneuvers or localization, which are among the most challenging aspects of autonomous racing. Moreover, we describe how we implemented a fault injection module, capable of introducing sensor delays and perturbations as well as modifying outputs of any node of the stack. Finally, we describe the design of our automated reporting process, aimed at maximizing the effectiveness of the simulation analysis. 
\end{abstract}

\section{INTRODUCTION}

In the last decade, autonomous vehicles have been ported from restricted and constantly supervised environments to the public roads without an on-board human supervisor. In particular, in San Francisco (USA), robotaxi services operates daily without fatalities and with very limited cases of severe injuries. Despite this, minor crashes and road impediments caused by robotaxis are increasing. With a lack of reporting and demonstration of improvement on these matters by the majority of the companies involved, the public sentiment and trust in the service is at risk \cite{cummings2024assessing}.
In \cite{waymo2018safety}, Waymo reported a safety case of its autonomous driving and its Operational Design Domain (ODD). A brief overview of the testing and validation methods is provided. In particular, they exploit simulation to repeat a certain scenario with different trajectories and positions of objects with a process called fuzzing. A description of their simulator, Waymax, is given in \cite{waymax}. They use the initial seconds of real data to initialize the objects and the environment of the simulation. Some metrics, such as collisions, off-road, and the feasibility of the vehicle's action, are used to assess the correctness of the software. In \cite{uber2018safety}, Uber produced a safety report mentioning an offline testing toolchain that perform regression tests, unit tests, and checks the correct integration between the software stack and the map used. 
None of the previously described reports mentions the definition and injection of faults in the simulation environment, such as objects moving erratically, delays, or malfunctions in the autonomous stack, but they mainly mention nominal driving conditions.

\begin{figure}[t]
  \centering
  \includegraphics[width=1.0\columnwidth]{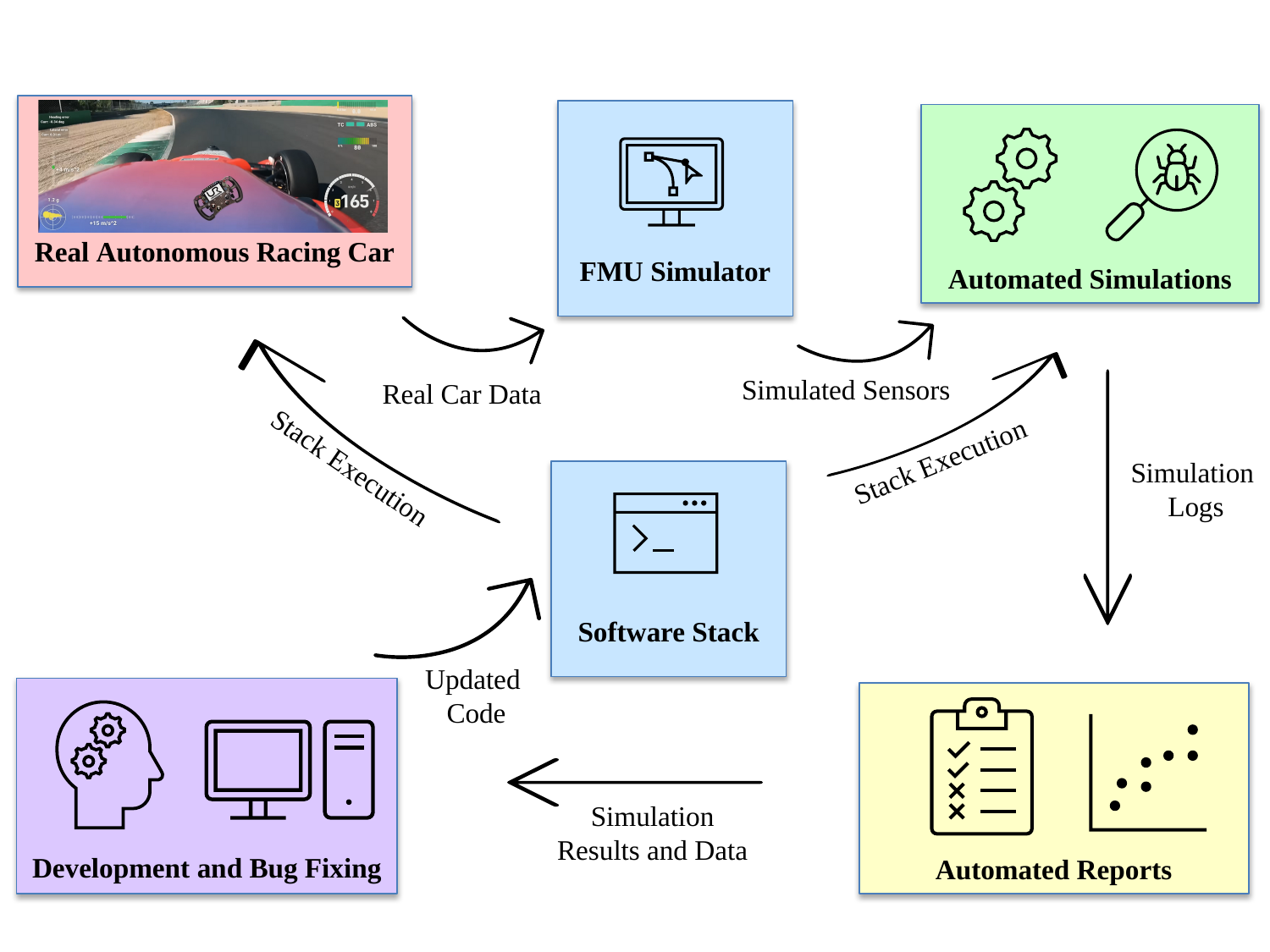}
  \caption{Development workflow scheme}
  \label{img:development_workflow}
\end{figure}

A research application of autonomous driving where simulation is extremely important is autonomous racing \cite{betz_survey}. In this domain, the software stack is pushed to its operational limits under extreme conditions that can differ significantly from what was expected. 
Although these events can occur generally in any situation, they are notably more common in this context, characterized by significant vibrations and mechanical strain. Moreover, even if human safety is not at stake, a crash may lead to damage worth hundreds of thousands of dollars, slowing down the research and making it financially unsustainable.
In the literature, several works describe the architecture of autonomous racing stacks and the safety mechanisms they implement. 
In \cite{paper_humda}, the scalable supervisory architecture of the Hungarian Mobility Development Agency (HUMDA) racing team is presented, focusing on modularity and redundancy. In ~\cite{paper_tum}, the TUM Autonomous Motorsport team, from Technical University of Munich (TUM), presents the software integration challenges they encountered developing their racing software stack and the safety procedures they follow. Cavalier Autonomous Racing, in \cite{paper_cavalier}, describes their architecture based on safety actions to mitigate faults, monitoring the health of the nodes and the data used.
None of these papers on autonomous racing describes the testing phase. Only in \cite{autosim_tum} the TUM's full software architecture detail its modular design and integration strategies, mentioning their automation testing pipeline. Although, their framework primarily targets high‑level validation of the entire stack and does not specifically focus on the efficient, reproducible testing of targeted scenarios such as overtaking maneuvers or faults. 
 
This paper wants to give a contribution in enhancing the effectiveness and efficiency of automated testing in safety-critical autonomous driving applications, demonstrating its features and potential in an extreme condition as the racing context. 
In particular, we describe how we managed to automate the testing and reporting for our autonomous racing stack, ur.autopilot \cite{er.autopilot.1.1}, developed to compete against other universities from all over the world, in two competitions: Indy Autonomous Challenge (IAC) \cite{iac} and Abu Dhabi Racing League (A2RL) \cite{a2rl}.
The stack uses a proprietary pub/sub framework, similar to Robot Operating System (ROS)\cite{ros2}.
The proposed approach allows for efficient evaluation of different scenarios, reproduced using different stack configurations, speeding up the execution of a high-fidelity simulator, and injecting faults into the system. This process aims to create an effective and continuous development and testing workflow, as depicted in Fig. \ref{img:development_workflow}.

The accurate simulation's physics, and in particular its initialization and integration into the testing pipeline, and its execution speedup are presented in Section \ref{sec:fmu_init}. In Section \ref{sec:auto_sim}, we describe the automatic simulation process based on configuration files and scripts. The module that injects faults into the simulation is presented in Section \ref{sec:fault_simulation}. The complete automated testing and reporting system, able to automatically start, finish, and analyze the simulation, is described in Section \ref{sec:automatic_reports} while its integration in the GitHub Pipeline is presented in Section \ref{sec:github_pipeline}. Examples of scenarios and the results are presented in Section \ref{sec:results}. The conclusions and future works are discussed in Section \ref{sec:conclusions}.



\section{FMU AND STACK INITIALIZATION}
\label{sec:fmu_init}

\subsection{FMU simulator}
A Functional Mock-up Unit (FMU) is a standardized software component that facilitates model exchange and co-simulation, as defined by the Functional Mock-up Interface (FMI) standard \cite{fmi_standard}. FMUs enable seamless integration of models developed in different simulation environments into a unified simulation framework. In this work, an FMU is used to represent a multibody vehicle model, allowing for accurate and high-fidelity simulation of the vehicle's dynamic behavior. The accuracy of the physics compared to real data is demonstrated in \cite{er.autopilot.1.1}. The sensors’ noise is also modeled to closely replicate real-world characteristics. Currently, all sensors can be replicated, with the exception of Light Detection and Ranging (LiDARs) and cameras.

To simulate realistic driving scenarios, the vehicle interacts with a road surface defined using the Curved Regular Grid (CRG) format. CRG is a widely adopted standard for road representation that describes the surface over a regular grid in curvilinear coordinates. This format enables precise modeling of road features such as elevation changes, banking, and surface irregularities, making it highly suitable for advanced vehicle dynamics simulations.
\subsection{FMU initialization}
To initialize the vehicle at a specific location, its position must be specified in Frenet coordinates, defined as:
\begin{itemize}
    \item s: the longitudinal position along the track’s reference line (typically the track centerline)
    \item d: the lateral offset from the reference line
\end{itemize}
In addition to s and d, the vehicle's initial yaw angle relative to the heading of the reference line at the corresponding location must also be provided. This ensures that the vehicle is not only positioned correctly on the surface but also properly oriented for the intended motion.
It is important to highlight that the reference line used to generate the CRG surface is fixed and does not adapt to any alternative driving trajectories. Hence, Frenet coordinates defined with respect to an arbitrary trajectory cannot be directly interpreted in the context of the CRG reference line. Misalignment between the trajectory and the reference line can lead to inaccurate positioning and orientation of the vehicle in simulation.
To resolve this discrepancy, a two-step process is applied:
\begin{enumerate}
    \item Frenet-to-Cartesian Conversion:
Given a point (s,d) defined with respect to a desired driving trajectory, the corresponding Cartesian coordinates (x,y) are computed.
\item Cartesian-to-Frenet Re-projection (CRG-based):
Using the Cartesian coordinates obtained in the first step, a new transformation is performed to compute the Frenet coordinates (s,d) relative to the CRG reference line. Similarly, the vehicle’s initial yaw angle is corrected by calculating the angular difference between the heading of the desired trajectory and that of the CRG reference line at the corresponding location.
\end{enumerate}
This ensures that the vehicle's initial state is accurately mapped onto the CRG-defined road surface, maintaining both positional and directional consistency, even when the desired trajectory deviates from the CRG reference line. 

\subsection{Integration into the Stack}
\label{sec:integration_problems}
All the nodes of the stack need to be already running when the FMU simulation starts. Therefore, after launching the stack, there is a short delay before the FMU simulator starts. Still, this is not sufficient to allow the simulations to start correctly. Each module of the software stack must be adapted to guarantee the correct integration with the automated simulations (see Section \ref{sec:auto_sim}).
The major issues encountered and solved to make the initialization as smooth as possible, even at high speeds, are listed below.

\subsubsection{Localization module}
    This module receives in input sensors data and outputs the estimated position and velocity of the car in the world. It needs about one second from the moment it receives sensors data to initialize correctly and start publishing its messages. This is due to its complex internal filters structure. So, it is not possible to initialize the car with a speed different from zero in this way.
    The solution we applied was to introduce a new node that acts as a multiplexer (muxer) between the localization module and the ground truth from the FMU simulator. So, we remapped the topics of the localization, adding a common prefix to them, and used this new muxer node to publish the localization data. In particular, the node publishes the ground truth from the FMU for the first three seconds, to give the localization module time to initialize, and then simply forwards the data it publishes. As a result, after initialization, the stack is equivalent to the one used on the real car.
\subsubsection{Safety module}
    It is responsible for collecting all errors from the other modules of the stack and sending a stop command to the car to stop safely if necessary. So, when the stack is launched, it will continuously ask to stop the car until all the nodes are initialized and have started publishing their messages. To allow the car to start with a high initial speed in simulation, we added a configuration, activated automatically when using automatic simulations, to ignore all the errors triggered by the stack in the first three seconds. 
\subsubsection{Mission module}
    It manages the behavior of the car. It expects the car to start in the pit lane, and it makes the stack follow a specific sequence of rules and actions (maximum speed, minimum safety distance from opponent, among others). The module was adapted to include an optional configuration enabling the car to be spawned directly on track, bypassing the standard startup sequence.
\subsubsection{Controller module}
    It manages the actuators of the car (throttle, brake, steering). A new configuration was added to initialize the longitudinal controller with the same gear as the FMU, to avoid sudden reactions of the controller due to unexpected values at the beginning of the simulation.

\subsection{Simulation Speedup}
\label{sec:fmu_speedup}
An important aspect of simulations is that they can take a considerable amount of time to execute. Every test can last several minutes, and it can be unsustainable to run several in real time. Therefore, the FMU has been sped up by introducing a configurable speed-up factor, allowing the simulation to run faster than real time while preserving coherence with the real-time behavior. To achieve this, the following modifications were required: 
\begin{enumerate}
    \item Scaling of the FMU simulator time step. It was multiplied by the speed-up factor. In this way, the simulation time flows faster.
    \item Making the stack time flow faster. In our framework we have a clock master that manages the timestamp of all the other nodes of the stack. It was sufficient to use the FMU simulator as clock master, and make it publish the timestamp faster, accordingly to the speed-up factor. The same thing can be achieved in ROS, setting use\_sim\_time parameter to true and publishing the simulation timestamps on the /clock topic.
    \item Scaling of the frequency of the FMU and all the stack nodes. All processes run at a frequency defined with respect to the system clock time. Therefore, to be consistent with the \textit{faster than real time} simulation, the frequency of all nodes must be scaled accordingly with the speed-up factor, before starting them.
    
\end{enumerate}
These improvements ensure that the accelerated simulation is equivalent to the real-time one, as long as the hardware is able to keep up with it.
We noticed that in our software stack, the localization module is the bottleneck for the speed up, due to its complex internal filters structure. So, we added the possibility to run the simulations using the ground truth instead of the localization module, to test all the other modules more efficiently.  
In Table \ref{tab:fmu_speedup}, the performance obtained on laptops (i9-14900HX, 32GB RAM) and workstations (i9-14900KF, 128GB RAM) using the proposed software stack are reported. No significant differences were observed in the resulting data (e.g., tracking errors, understeer degree, lap time, etc.) compared to the real-time simulation. Speeding up further causes both the FMU and the stack itself to lag behind the requested frequency, and the simulations become inconsistent. 

\begin{table}[h]
\centering
\caption{FMU Speedup Comparison}
\begin{tabular}{lcc}
\hline
\textbf{Hardware} & \textbf{Loc Module} & \textbf{Ground Truth} \\
\hline
Laptop & 2x & 3x \\
Workstation & 3x & 4\textbf{}x \\
\hline
\end{tabular}
\label{tab:fmu_speedup}
\end{table}

\section{AUTOMATIC SIMULATIONS}
\label{sec:auto_sim}
The purpose of automatic simulations is to test many different scenarios in an efficient way, to find possible bugs in the autonomous driving software stack, and to validate new algorithms.

The process of creating and executing automatic simulations is relatively straightforward. Every simulation scenario is composed of two YAML files:
\begin{itemize}
    \item Configuration file: structured into two YAML documents. 
    \begin{itemize}
        \item The first one contains all the modified initial configurations of the stack. Both nodes parameters and simulation configurations (FMU initialization and speed-up factor) can be set.
        \item The second one contains the parameters for the automatic reports described in Section \ref{sec:automatic_reports}.
    \end{itemize}
    \item Scenario file: it contains a list of commands and configurations to be sent to the car during the simulation. Each group contains:
    \begin{itemize}
        \item lap: the lap during which the parameters should be sent
        \item s: longitudinal position in curvilinear coordinates, also referred to as Frenet coordinates, of the specified lap to send the parameters
        \item parameters: list of configurations and commands
    \end{itemize}
\end{itemize}

The automatic simulation script creates a copy of each configuration file of the stack, replacing all the parameters that are present in the scenario file to the corresponding node's configuration file. Then, a dedicated node of the stack parses the yaml scenario at startup and sends commands and configurations to the other nodes of the stack at the appropriate time defined in the scenario itself.
To stop the simulation, another dedicated node (launched only in simulation) checks whether the scenario is finished (the scenario manager node stopped publishing its heartbeat) or if an error was triggered by the stack. When one of these conditions is satisfied, the node finishes its execution, and the automatic simulation script stops the entire stack.

\section{FAULT SIMULATION}
\label{sec:fault_simulation}
This setup allows us to replicate the car's behavior in multiple scenarios and conditions. In order to take advantage of this, we implemented a module capable of introducing perturbations in the generated data, and we made them injectable at runtime. This method of action, combined with the extremely configurable and generalized fault definition, allows the fabrication of specialized faults that can simulate any issue that could be encountered in the real world.

\subsection{Module implementation}
The module is designed to exploit the nature of ROS-like frameworks, injecting the faults directly into the nodes’ communication by sending messages on the corresponding topic. A general configuration YAML file specifies the topics that the module should act upon by remapping them with a common suffix, ensuring that all other nodes publish their data to these new topics. The module then subscribes to all remapped topics and, by default, transparently republishes each message onto the original (unsuffixed) ones. In this way, it remains transparent to the rest of the stack, as other nodes continue publishing and subscribing without awareness that an intermediate module is present. When a specific topic fault configuration is set, it proceeds to simulate what is specified only on the section of the messages that needs it.

\subsection{Faults configuration}
The configuration file handles the definition of the faults to apply to each topic. For each node and each topic, a set of fault types is available, each one independent of the other:
\begin{itemize}
    \item Delay: essential to simulate latencies in data collection and processing. In real-time applications such as ours, delays on acquisition timestamps and inter-process data transmission can offset measures and severely disrupt strongly time dependent algorithms.
    \item Value multiplier: useful to simulate value stuck at 0 or unit conversion issues. 
    \item Value offset: basic introduction of constant offset to the message value. Useful to simulate sensor drift or biases. For example, Inertial Measurement Units (IMUs) can initialize with significant measurement biases.
    \item Value repetition: composed of the \textit{actual value} and a \textit{counter}, it overrides the data on the specified message field with the actual value (if specified, the current value otherwise) for N consecutive messages, with N equal to the value set in \textit{counter}.
    \item Gaussian noise: allows the simulation of random disturbances. The noise is characterized by a configurable mean and variance, enabling the reproduction of different operating conditions, from small fluctuations to severe disturbances. This approach is particularly useful for testing the robustness of localization and control algorithms against noisy inputs.
\end{itemize}

An example of a topic fault definition is the following, related to faults on the odometry:
\begin{lstlisting}[language=YAML, caption={Example configuration YAML for fault injection}, label={lst:yaml_config}]
localization: # stack module
  - name: /loc/odom # topic
    
    # [ms] (-1 to stop publication)
    delay: 20
    
    # fault definition per message (sub)field
    field_faults:
      pose:
        position: {fault_mult: 100}
        covariance: {
          fault_mult: 2,
          fault_repeat: {
            count: 100,
            value: 1000
          }
        }
      twist:
        covariance: {fault_mult: 0}
\end{lstlisting}

Using this approach, it is possible to simulate most of the issues encountered during practice in the field with the actual hardware, as it will be presented in Section \ref{sec:results}. Moreover, the system is designed to be easily extendable, allowing for the addition of new fault types and implementations if needed.

\subsection{Integration with automatic simulations}
The module, while capable of operating standalone or under manual control, proves especially powerful when integrated into the automatic simulation workflow. It exposes a custom service that makes the fault configuration editable at runtime, and through the use of the simulation scenario (see Section \ref{sec:auto_sim}), it is possible to create custom, space/time dependent fault applications across one or more simulation runs.

\section{AUTOMATIC TESTS AND REPORTS}
\label{sec:automatic_reports}
Automatic simulations are an effective mean of significantly reducing testing time.
However, for them to be effective, the evaluation of the results must also be automated. Thus, we implemented another pipeline that, when an automatic simulation finishes its execution, analyzes its logs. In particular, there are three steps to produce the reports: data extraction and processing, tests' execution, and generation of final reports. Each step was designed to be modular, flexible, and scalable. The software stack evolves very rapidly, and it must be easy to adapt to changes. Moreover, our solution could also be used with other autonomous driving stacks' logs, changing almost exclusively configurations. 

\subsection{Data extraction}
The first step is data extraction. From the simulation logs, a CSV file is generated for each topic. Each of them contains multiple columns. The first one is always the timestamp; then there is a column for each variable published on that topic. Each file is parsed into a tabular structure using pandas\cite{pandas}, a Python module. Subsequently, the extracted tables are merged together to create a single common table. This is needed to correlate data that are published on different topics. To achieve this, the timestamps of each table are analyzed to determine the maximum frequency, which is then used to establish a common timestamp reference for all tables. Each individual table is then merged into the master table using a nearest-neighbor interpolation technique.

In order to avoid having too much data to analyze: 
\begin{itemize}
\item There is a maximum bound on the common frequency to avoid having too much data to perform the following computations. 
\item Only data that are effectively used for tests or reports are merged into the global reference table.
\end{itemize}

\subsection{Data processing}
In this step, the data are cleaned and aggregated. They are divided into laps, and relevant information is extracted for subsequent analysis. For each lap, statistical metrics such as the mean and maximum path tracking error, the yaw rate, and the understeer degree of the vehicle are calculated.
The mapping between each raw variable used in the code and its corresponding topic is defined in a dedicated YAML file, allowing easy updates to topic or sub-topic names without modifying the code.
\subsection{Tests}
There is a list of tests that can be executed, some are similar to the one used in \cite{waymax}, with racing-specific checks. It is modular and can be easily extended by adding new tests. The ones already implemented in our software stack are:

\begin{itemize}
    \item Tracking errors: checks if the car exceeded the safety lateral and/or heading thresholds related to the ability of the controller to follow a path generated by the planner.
    \item Car started: checks the total length of the track covered, ensuring that both the stack and the FMU simulator have actually started.
    \item Car stopped: checks if the car stopped unexpectedly without a specific command from the automatic simulation scenario.
    \item Stack errors: checks whether any errors were triggered by the stack's nodes and, if so, whether the vehicle was able to stop safely on the track.
    \item Track boundaries: checks if the car went out of track.
    \item Vehicle dynamics metrics: checks if the following quantities are above their configurable safety threshold: understeer/oversteer degree, sideslip angle, yaw rate, and lateral velocity.
    \item Ghost opponents collisions: checks if, during the simulation, the ego vehicle collided with an opponent. Collisions are not simulated in the stack; the opponents are only "ghosts". Thus, there is not direct feedback in real time to the ego car. The collision check is done by analyzing the logs after the simulation has finished, comparing the ground truth position of the ego vehicle with those of the ghost vehicles.
\end{itemize}

For all the tests that require the position of the ego car, the ground truth of the FMU simulator is used, instead of the one estimated by the localization module. In fact, the localization may produce inaccurate estimates of the vehicle’s position.
By default, all the test are selected, but it is possible to decide, for every scenario, to exclude one or more of them. For example, while testing sensor faults, detecting generic stack nodes errors could be redundant. In this case, it is better to exclude all the tests that detect the errors coming from the other nodes, considering only the one that checks the distance from the track boundaries, verifying that the car managed to stop safely on the track after detecting the delays. 
For each simulation run, the script generates a JSON report containing general information (e.g., best lap time) and a list of errors that occurred. Each error also includes details about where it occurred (lap and Frenet coordinates). If the simulation was successful, the error list will be empty. This report is then used by the GitHub pipeline (see Section \ref{sec:github_pipeline}).

\subsection{Final reports}
Two types of reports are generated at the end of the tests: One contains more general information, and the other contains module-specific information.
\subsubsection{General run overview}
\label{sec:overview_report}
Overview of some basic information regarding the run. It contains:
\begin{itemize}
    \item Lap info table: for each lap, information about lap time, maximum, and average speed
    \item Errors table: list of occurred errors (if any) with error description and information on where it happened (lap and Frenet coordinates)
    \item Vehicle dynamics metrics: a table for understeer/oversteer degree, sideslip angle, yaw rate, and lateral velocity that shows their maximum and average values for each lap. Possibly dangerous values are highlighted using colors. Two threshold are implemented: a yellow warning one and a red error one (this represents that the simulation failed). All thresholds have default values, but can be fine tuned for every scenario.
    \item Ghost opponents info: this is added to the report only if the scenario involves a ghost opponent (to test interactions). It is composed by:
    \begin{itemize}
        \item Collisions with opponents: for each collision, a schematic plot representing the incident is generated, showing the vehicles' positions and orientations.
        \item Overtakes table: for each ghost opponent, it is checked whether the ego vehicle went from being behind to ahead of it, to detect the overtakes. Then, for every overtake, some information is added:
        \begin{itemize}
            \item Success or collision
            \item Lap and Frenet coordinates at the start and end of the overtake (an overtake is considered started when the ego vehicle approaches the opponent at -30\,m, and ended when it is ahead by +20\,m)
            \item Time to overtake
            \item Average delta speed between ego and opponent during the overtake
        \end{itemize}
    \end{itemize}
    
\end{itemize} 

\subsubsection{Module-specific reports}
\label{sec:specific_reports}
Reports are generated for specific modules to monitor their behavior and performance. These reports include data unrelated to the simulated environment, such as sensors performance data, as they are also used to analyze real test-run data. The reports are as follows:
\begin{itemize}
    \item Localization: includes graphs and data on the actual vehicle trajectory, sensor performance (IMU and GPS messages delays, GPS signal quality including satellite count, and measurement covariance), and the covariance of the estimated vehicle state during the run.
    \item Control: includes graphs and data on actuation command tracking (sent versus feedback), tracking error relative to the generated trajectory, speed tracking error, and path-planning status.
    \item Perception: includes graphs and data on opponent detection and tracking, perception pipeline execution times, and performance.
\end{itemize}

\section{GITHUB PIPELINE INTEGRATION}
\label{sec:github_pipeline}
\begin{figure}[!t]
  \centering
  \hspace*{0.25cm} 
  \includegraphics[width=1.0\columnwidth]{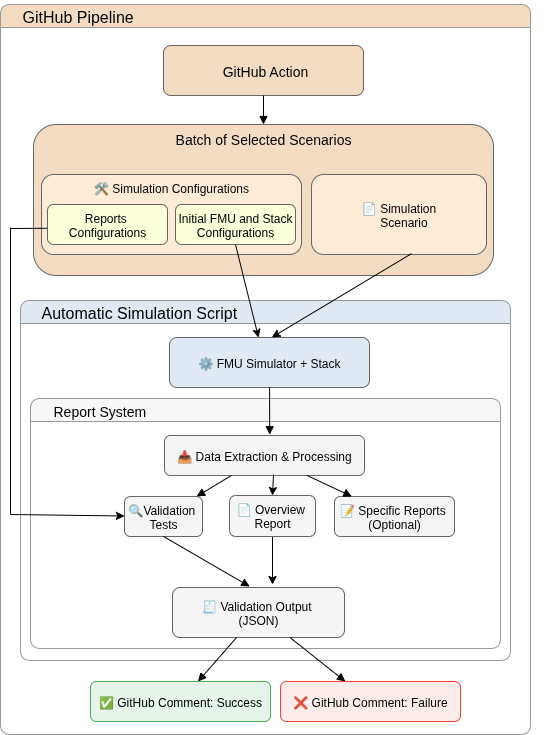}
  \caption{Overview of the automatic simulation workflow.}
  \label{img:autosim_pipeline}
\end{figure}

Every time a new GitHub Pull Request (PR) is created, an automatic simulation pipeline is triggered. An overview of the complete pipeline is given in Fig. \ref{img:autosim_pipeline}.\\
This includes:
\begin{itemize}
    \item Clang code format checks
    \item Compilation of the code
    \item Execution of automatic simulations and reports
\end{itemize}

For this last point, there is a basic performance scenario that is executed for every PR, to test that the new code is not breaking anything (causing the stack to trigger some errors or the car to crash) and is not affecting the overall performance. Moreover, based on specific tags, targeted tests can be executed. For example, if the PR is modifying the code for the overtakes, there are a batch of overtake scenarios that are executed. 

If at least one of the simulations executed fails one or more tests, the whole pipeline will fail. In this case, it is straightforward to reproduce the problem, executing locally the simulation that failed. When the problem is solved, pushing a new commit on the branch, the GitHub pipeline will be triggered again automatically.

When executing automatic simulations, only the overview report is produced(see Section \ref{sec:overview_report}) to reduce the computation time. Then, if the pipeline fails, it is possible to store the simulation logs. This allows, if necessary, the a posteriori generation of the report for the module that caused the pipeline to fail or the direct manual analysis of the logs.
\begin{figure*}[htbp]
    \centering
    \begin{subfigure}[t]{0.48\textwidth}
        \centering
        \includegraphics[width=\linewidth]{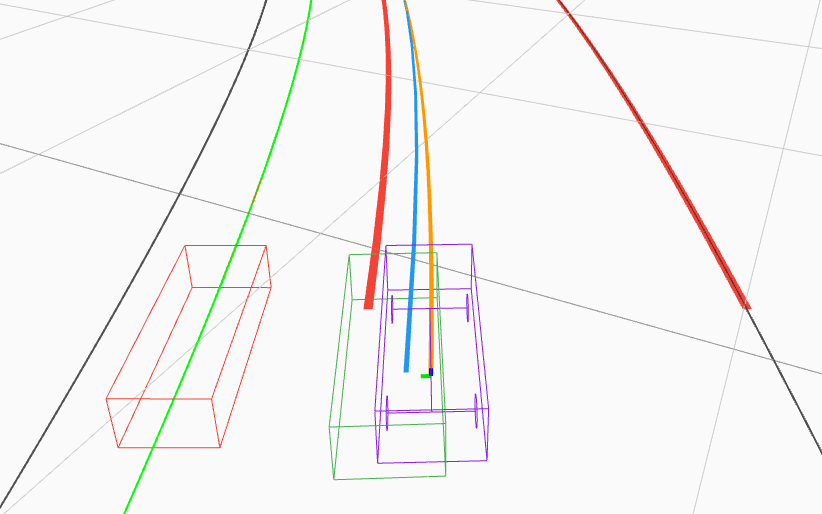}
        \caption{Successful overtake despite the fault injection}
        \label{img:fault_slam_100ms}
    \end{subfigure}%
    \hfill
    \begin{subfigure}[t]{0.48\textwidth}
        \centering
        \includegraphics[width=\linewidth]{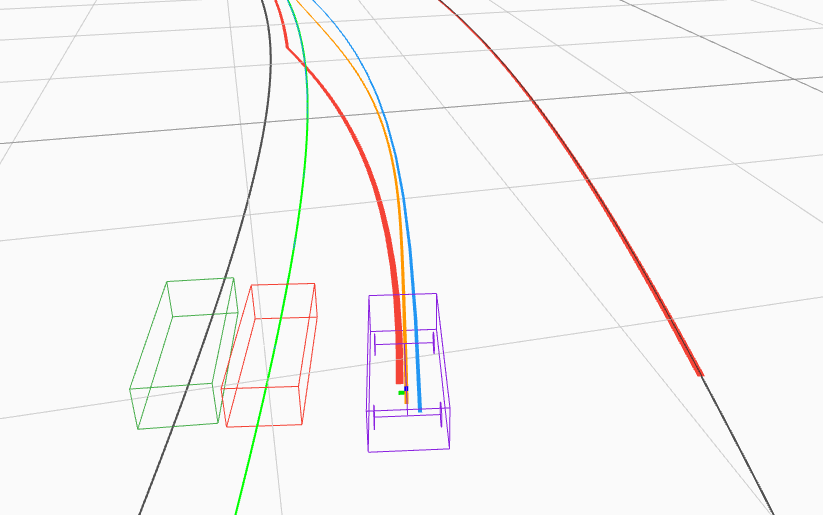}
        \caption{Collision with ghost and track boundaries during the overtake maneuver}
        \label{img:fault_slam_500ms}
    \end{subfigure}
    \caption{Two different faults injected in the same scenario.}
    \label{fig:overtake_fault_combined}
\end{figure*}
\section{RESULTS}
\label{sec:results}
In this section, we present some examples of scenarios that target some of the most critical modules and help assure and improve the safety of the stack. It is always possible to decide which of them to execute in the pipelines (for example overtake scenarios could be excluded if the testing regards only the ego vehicle without interaction with any agent).

\subsection{Single performance scenario}
The main goal of racing is the minimization of lap time. For each track there is a scenario that emulates the dynamics of a qualifying race progression. It contains:
\begin{itemize}
    \item Racing line initialization: among the set of trajectories computed offline, the optimal one to minimize lap time is chosen.
    \item Performance increment: at the beginning of each lap, the performance cap is increased as we would do on track.
\end{itemize}
In a single simulation, it is possible to test the general behavior of the stack, the pit exit, the pit entry, and the performance, to verify that the whole stack is working (in the nominal case scenario). Then it is also possible to check the performance (the best lap time is always printed in a GitHub comment)

\subsection{Ghost overtake scenarios}
In a multi-car race, the ability to interact with opponents is also essential. 
Thanks to the FMU initialization, one can create, for each track, different overtake scenarios, with one or multiple opponents. The tests can cover both attacking and defensive maneuvers. In fact, when being overtaken, the car needs to follow strict competition rules to avoid collisions. 



\subsection{Fault simulation scenarios}
The safety and robustness of the stack are crucial points to consider in order to reduce the risk of incidents as much as possible, keeping the research project sustainable. 
Three types of safety scenario tests are implemented.
\subsubsection{Minor faults}
They include faults that the software stack should handle without problems. For example, small delays or outliers on sensors data. 
\subsubsection{Severe faults}
They include mainly huge delays in sensors to test the robustness of the localization or control module (see Figure \ref{fig:overtake_fault_combined}). The complete loss of specific nodes is also tested, such as the controller (the stack can stop the vehicle switching to a backup controller). For these types of scenarios, it is convenient to disable all tests except the one on track boundaries, to verify whether the vehicle is able 
to stop safely in different scenarios (speed and track position). This is useful to fine tune the safety thresholds for the tests and racing events on track.

\subsubsection{Unrecoverable faults}
They test known single points of failure in the stack, which, even if considered extreme and improbable, they should be managed. One example is the crash of the localization node. In these cases, the only thing that can be done is to apply high pressure on the brakes, keeping the steering straight. The test is successful if the vehicle correctly triggers the emergency procedure and is able to stop without going off track in low-complexity scenarios (on the straight at moderate speed).

\subsection{Scenario example - overtake with fault}

In this section we provide an example of how executing the same scenario with slightly different parameters helps to efficiently test the robustness of the stack.
In this example scenario, in the Las Vegas Motor Speedway (LVMS) oval track, the ego vehicle is initialized at 270 km/h, and a ghost opponent is initialized at 220 km/h, 100m in front of it. The scenario consists of the injection of a fault at the beginning of the overtake. 

In Figs \ref{fig:overtake_fault_combined}, there is a representation of the moment of overtake with two different faults.
The rectangular boxes represent the cars position on the track:
\begin{itemize}
    \item purple: localization estimated ego position
    \item green: FMU ground truth ego position
    \item red: ghost opponent ground truth position
\end{itemize}
The green line plots the offline computed racing line that the vehicle tries to track, the red one represents the tunnel created by the planner to overtake the opponent safely, the blue is the planner trajectory, and the orange is the controller's one.

In Fig \ref{img:fault_slam_100ms}, the injected fault is the simultaneous complete loss of all GPS and Optical sensor measurements, and a constant 100 ms delay on all the IMUs and on the LiDAR Inertial Odometry (LIO) state estimation. The car oscillates slightly, but it is still able to perform the overtaking maneuver without colliding and then stop safely, giving space to the opponent while slowing down.

In Fig \ref{img:fault_slam_500ms}, the delay on the LIO is increased to 500 ms. The localization module fails to determine the vehicle’s position, and the car crashes into the opponent before going off track.

Multiple simulations of the same scenario can be executed, with the fault severity progressively increased, to determine the maximum level of resilience before the vehicle crashes. The batch of scenarios can also be executed with different stack configurations to efficiently understand if there is an improvement. The same concept can be used to improve lap time, overtakes, etc. 

\subsection{Real world scenario}
\begin{figure}[!h]
    \centering
    \begin{subfigure}[t]{0.45\columnwidth}
        \centering
        \includegraphics[width=\linewidth]{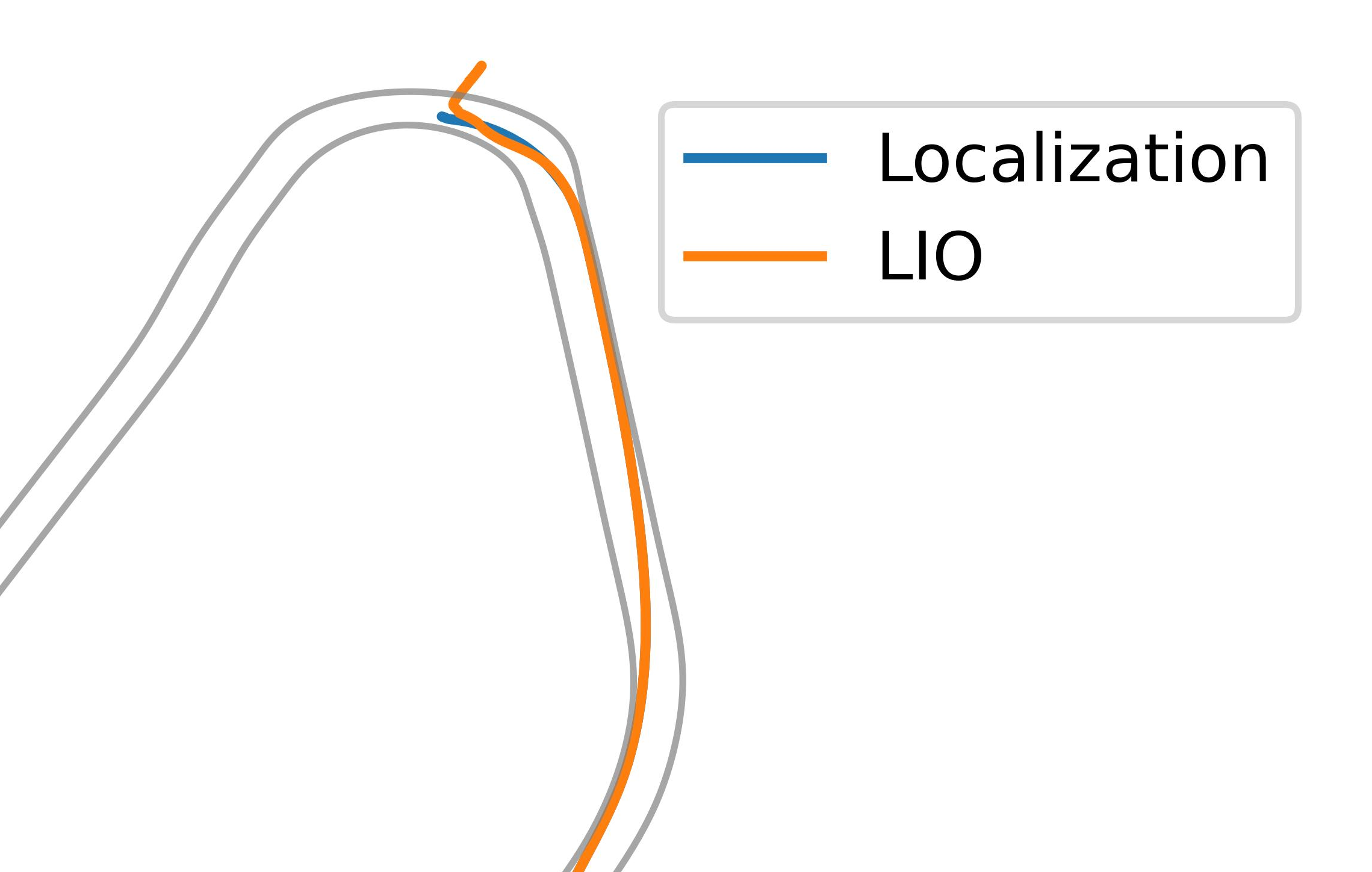}
        \caption{LIO estimation off track (real data)}
        \label{img:lio_vs_loc}
    \end{subfigure}%
    \hfill
    \begin{subfigure}[t]{0.45\columnwidth}
        \centering
        \includegraphics[width=\linewidth]{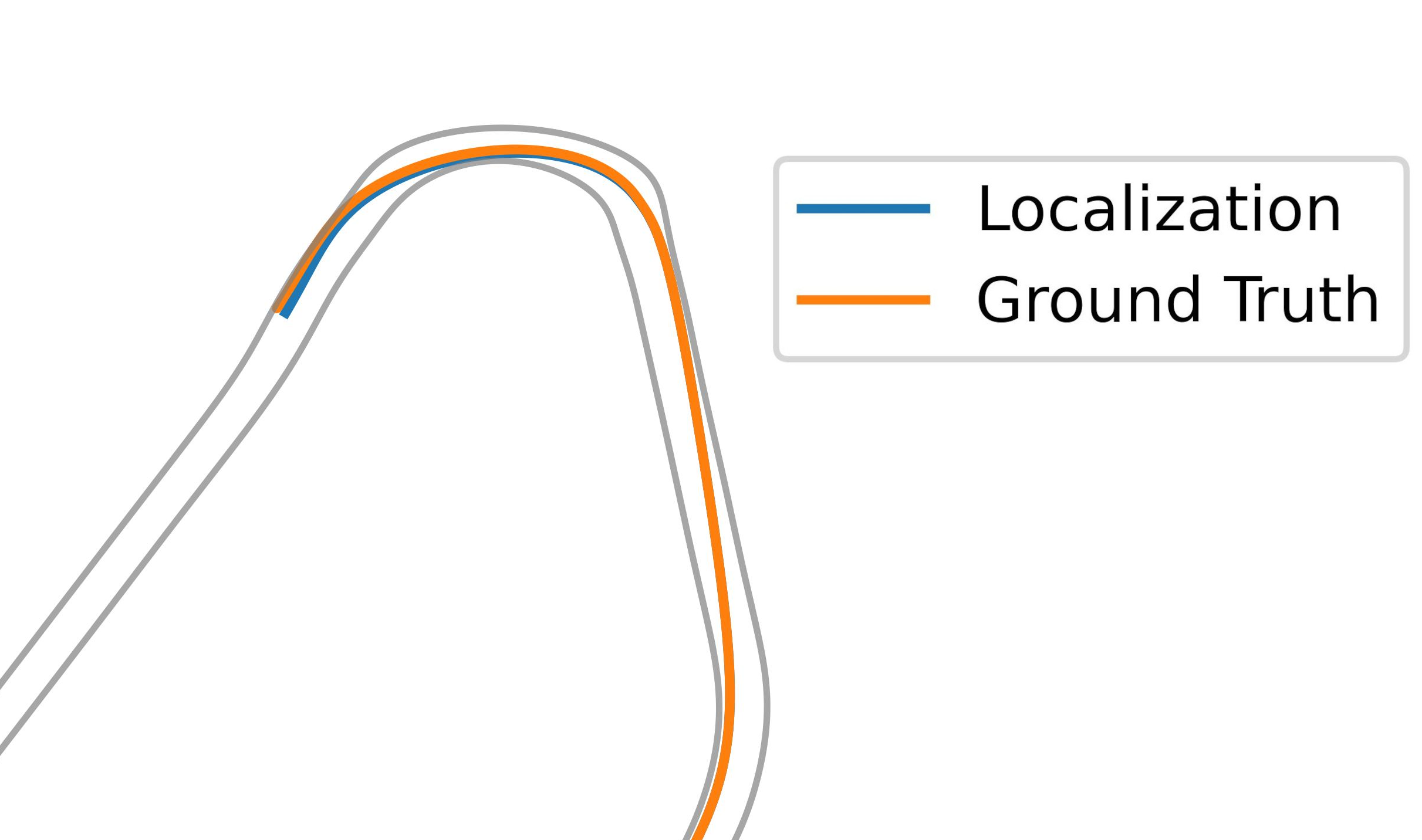}
        \caption{Localization estimation and ground truth (sim data)}
        \label{img:loc_vs_gt}
    \end{subfigure}
    \caption{IMU delay fault, reality vs simulation comparison}
    \label{fig:real_vs_sim}
\end{figure}

\begin{figure}[!h]
    \centering
        \centering
        \includegraphics[width=\linewidth]{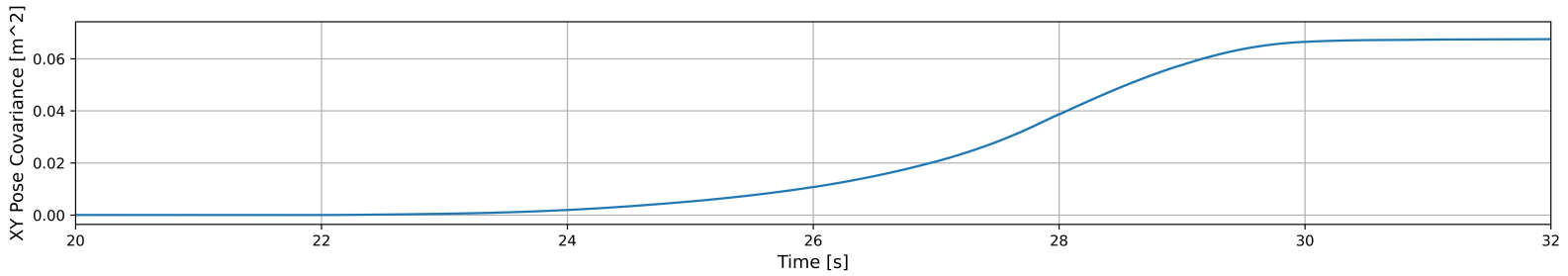}

        \includegraphics[width=\linewidth]{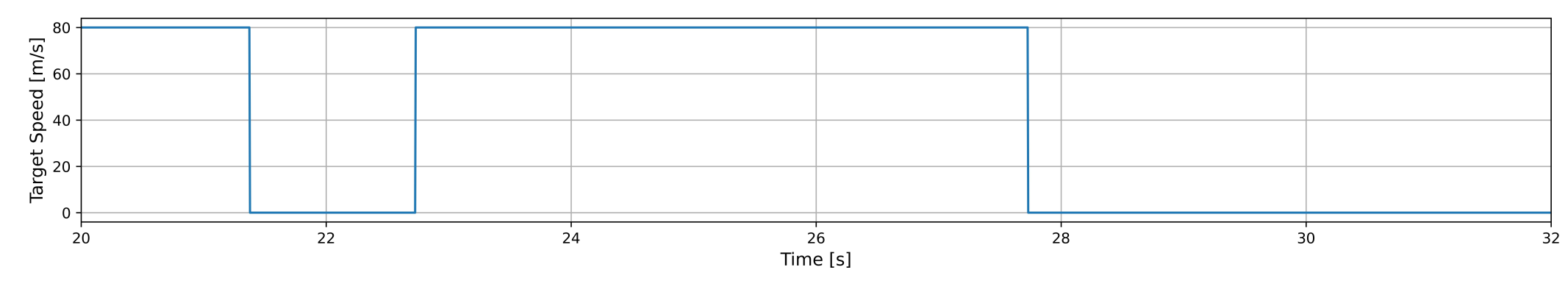}
        \caption{Target speed is set to zero when localization position estimation exceeds $0.03\ \mathrm{m}^2$}
        \label{img:sim_imu_delay_errrors}
\end{figure}

It is also possible to create scenarios that replicate the faults that occurred in the real world. An example is the scenario that has been developed to reproduce the A2RL race event on April 27, 2024, at the Yas Marina Formula 1 circuit. Unimore's car experienced a severe delay in IMUs and GPS entering Turn 5. Although GPS are not used in the Yas Marina Circuit (due to their poor quality signal), IMUs are essential. A 400 ms drop of the IMUs measurements made the LIO estimation go off track, as shown in Fig. \ref{img:lio_vs_loc}. The software stack detected it and was able to ban that source, relying only on the model to stop safely. Thanks to the simulation scenario, it is possible to replicate the problem and test what would have happened if the car hadn't stopped. In this case, the car would have been able to continue without LIO for a while, deviating only slightly from the ground truth, until the localization covariance exceeded the safety threshold of $0.03\ \mathrm{m}^2$, leading to a safety stop, as shown in Fig \ref{img:sim_imu_delay_errrors}. Therefore, if the slam could re-initialize before the covariance hit the safety threshold, it would have been possible to continue the race without stopping. Testing these types of conditions in simulation is essential to understand the software stack's limits. This is fundamental both to improve it adding new features and to assess the risks of increasing the safety thresholds for the race event, in order to find the best trade-off between losing a race for an unnecessary safety stop and risking a crash.

\section{CONCLUSIONS}
\label{sec:conclusions}
Automatic simulations are now actively used in the GitHub pipelines of the team as an important step before merging new PRs. Especially when there is limited time to test new features before the car goes on track, an automatic pipeline that can detect problems with last-minute configurations is fundamental. It allows the team to deploy new improvements of the software stack faster and reliably. The pipeline is automatically executed on the track the team is working on for the most imminent event. Continuous testing of several slightly different scenarios is also beneficial to identify possible hard-to-reproduce bugs, such as those related to the concurrency of the resources used by the stack.

Future work will include the addition of LiDAR sensor data to simulate the LIO algorithm properly. A further feature to improve the realism of the simulation is to replicate the low-level CAN and Ethernet interfaces in order to validate the correctness of the sensors' drivers included in the stack. Large Language Models (LLMs) will be explored to produce a detailed but effective description of the reports produced by the pipeline presented in this paper.
\addtolength{\textheight}{-12cm}   




\section*{ACKNOWLEDGMENT}
We would like to thank the entire Unimore Racing team for their support, collaboration, and valuable contributions throughout this work.

\bibliographystyle{unsrt}   
\bibliography{references}   

\end{document}